%% file: main.tex
\documentclass{article} 

\usepackage[T1]{fontenc}
\usepackage[colorinlistoftodos]{todonotes}
\usepackage[english]{babel}
\usepackage[font=small,labelfont=bf]{caption}
\usepackage[ruled, vlined]{algorithm2e}
\usepackage[utf8x]{inputenc}
\usepackage{amsfonts}
\usepackage{amsmath}
\usepackage{amssymb}
\usepackage{bbm}
\usepackage{bm}
\usepackage{booktabs}
\usepackage{color}
\usepackage{graphicx}
\usepackage{iclr2018_conference,times}
\usepackage{tabularx}
\usepackage{url}
\usepackage{wrapfig}
\usepackage{sidecap}
\usepackage{rotating}
\usepackage{multirow}
\usepackage[colorlinks=true, citecolor=black, urlcolor=black]{hyperref}
\usepackage{multicol}

\iclrfinalcopy

\newcommand*\samethanks[1][\value{footnote}]{\footnotemark[#1]}

\DeclareMathOperator*{\argmax}{argmax}
\DeclareMathOperator*{\var}{var}
\DeclareMathOperator*{\skewness}{skew}
\DeclareMathOperator*{\kurt}{kurt}

\let\tb\textbf
\title{Meta-Learning for Semi-Supervised Few-Shot Classification}

\author{
Mengye Ren${}^{\dag\Join}$, 
Eleni Triantafillou\thanks{Equal contribution.}${\ \ }^{\dag\Join}$, 
Sachin Ravi\samethanks[1]${\ \ }^\S$, 
Jake Snell$^{\dag\Join}$, 
Kevin Swersky$^\P$, \\
\textbf{Joshua B. Tenenbaum$^\natural$, 
Hugo Larochelle${}^{\P\ddagger}$ \& 
Richard S. Zemel${}^{\dag\ddagger\Join}$}\\
${}^\dag$University of Toronto, 
${}^\S$Princeton University, 
${}^\P$Google Brain, 
${}^\natural$MIT, 
${}^\ddagger$CIFAR, 
${}^{\Join}$Vector Institute \\
\texttt{\{mren,eleni\}@cs.toronto.edu, sachinr@cs.princeton.edu},\\
\texttt{jsnell@cs.toronto.edu, kswersky@google.com},\\
\texttt{jbt@mit.edu, hugolarochelle@google.com, zemel@cs.toronto.edu}
}

\begin{document}
\maketitle

\input{sections/abstract}
\input{sections/introduction}
\input{sections/background}
\input{sections/model}
\input{sections/related}
\input{sections/experiments}

\input{sections/conclusion}
\input{sections/acknowledgement}

\bibliography{references}
\bibliographystyle{iclr2018_conference}

\newpage
\appendix
\input{sections/supplementary}

\end{document}

%% file: sections/abstract.tex
\begin{abstract}
In few-shot classification, we are interested in learning algorithms that train a classifier from
only a handful of labeled examples. Recent progress in few-shot classification has featured 
meta-learning, in which a parameterized model for a learning algorithm is defined and trained on episodes
representing different classification problems, each with a small labeled training set and its
corresponding test set. In this work, we advance this few-shot classification paradigm towards a
scenario where unlabeled examples are also available within each episode. We consider two
situations: one where all unlabeled examples are assumed to belong to the same set of classes as the
labeled examples of the episode, as well as the more challenging situation where examples from other
{\it distractor} classes are also provided. To address this paradigm, we propose novel extensions of
Prototypical Networks~\citep{snell2017protonet} that are augmented with the ability to use unlabeled
examples when producing prototypes. These models are trained in an end-to-end way on episodes, to
learn to leverage the unlabeled examples successfully. We evaluate these methods on versions of the
Omniglot and {\it mini}ImageNet benchmarks, adapted to this new framework augmented with unlabeled
examples. We also propose a new split of ImageNet, consisting of a large set of classes, with a
hierarchical structure. Our experiments confirm that our Prototypical Networks can learn to improve
their predictions due to unlabeled examples, much like a semi-supervised algorithm would.
\end{abstract}

%% file: sections/introduction.tex
\section{Introduction}

The availability of large quantities of labeled data has enabled deep learning methods to achieve
impressive breakthroughs in several tasks related to artificial intelligence, such as speech
recognition, object recognition and machine translation. However, current deep learning approaches
struggle in tackling problems for which labeled data are scarce. Specifically, while current methods
excel at tackling a single problem with lots of labeled data, methods that can simultaneously solve
a large variety of problems that each have only a few labels are lacking. Humans on the other hand
are readily able to rapidly learn new classes, such as new types of fruit when we visit a tropical
country. This significant gap between human and machine learning provides fertile ground for deep
learning developments.


For this reason, recently there has been an increasing body of work on few-shot learning, which
considers the design of learning algorithms that specifically allow for better generalization on
problems with small labeled training sets. Here we focus on the case of few-shot classification,
where the given classification problem is assumed to contain only a handful of labeled examples per
class. One approach to few-shot learning follows a form of meta-learning \footnote{See the following
blog post for an overview: \url{http://bair.berkeley.edu/blog/2017/07/18/learning-to-learn/}}
~\citep{Thrun1998, Hochreiter2001}, which performs transfer learning from a pool of various
classification problems generated from large quantities of available labeled data, to new
classification problems from classes unseen at training time. Meta-learning may take the form of
learning a shared metric~\citep{vinyals2016matchingnet,snell2017protonet}, a common initialization
for few-shot classifiers~\citep{ravi2017oneshot,FinnC2017} or a generic inference
network~\citep{Santoro2016,MishraN2017}.

\input figures/motivation_figure

These various meta-learning formulations have led to significant progress recently in few-shot
classification. However, this progress has been limited in the setup of each few-shot learning
episode, which differs from how humans learn new concepts in many dimensions. In this paper we aim
to generalize the setup in two ways. First, we consider a scenario where the new classes are learned
in the presence of additional unlabeled data. While there have been many successful applications of
semi-supervised learning to the regular setting of a single classification
task~\citep{ChapelleO2010} where classes at training and test time are the same, such work has not
addressed the challenge of performing transfer to new classes never seen at training time, which we
consider here. Second, we consider the situation where the new classes to be learned are not viewed
in isolation. Instead, many of the unlabeled examples are from different classes; the presence of
such {\it distractor} classes introduces an additional and more realistic level of difficulty to the
few-shot problem.

This work is a first study of this challenging semi-supervised form of few-shot learning. First, we
define the problem and propose benchmarks for evaluation that are adapted from the Omniglot and {\it
mini}ImageNet benchmarks used in ordinary few-shot learning. We perform an extensive empirical
investigation of the two settings mentioned above, with and without distractor classes. Second, we
propose and study three novel extensions of Prototypical Networks~\citep{snell2017protonet}, a
state-of-the-art approach to few-shot learning, to the semi-supervised setting. Finally, we
demonstrate in our experiments that our semi-supervised variants successfully learn to leverage
unlabeled examples and outperform purely supervised Prototypical Networks.

%% file: figures/motivation_figure.tex
\begin{wrapfigure}{r}{0.5\textwidth}
    \centering
    \includegraphics[width=0.48\textwidth]{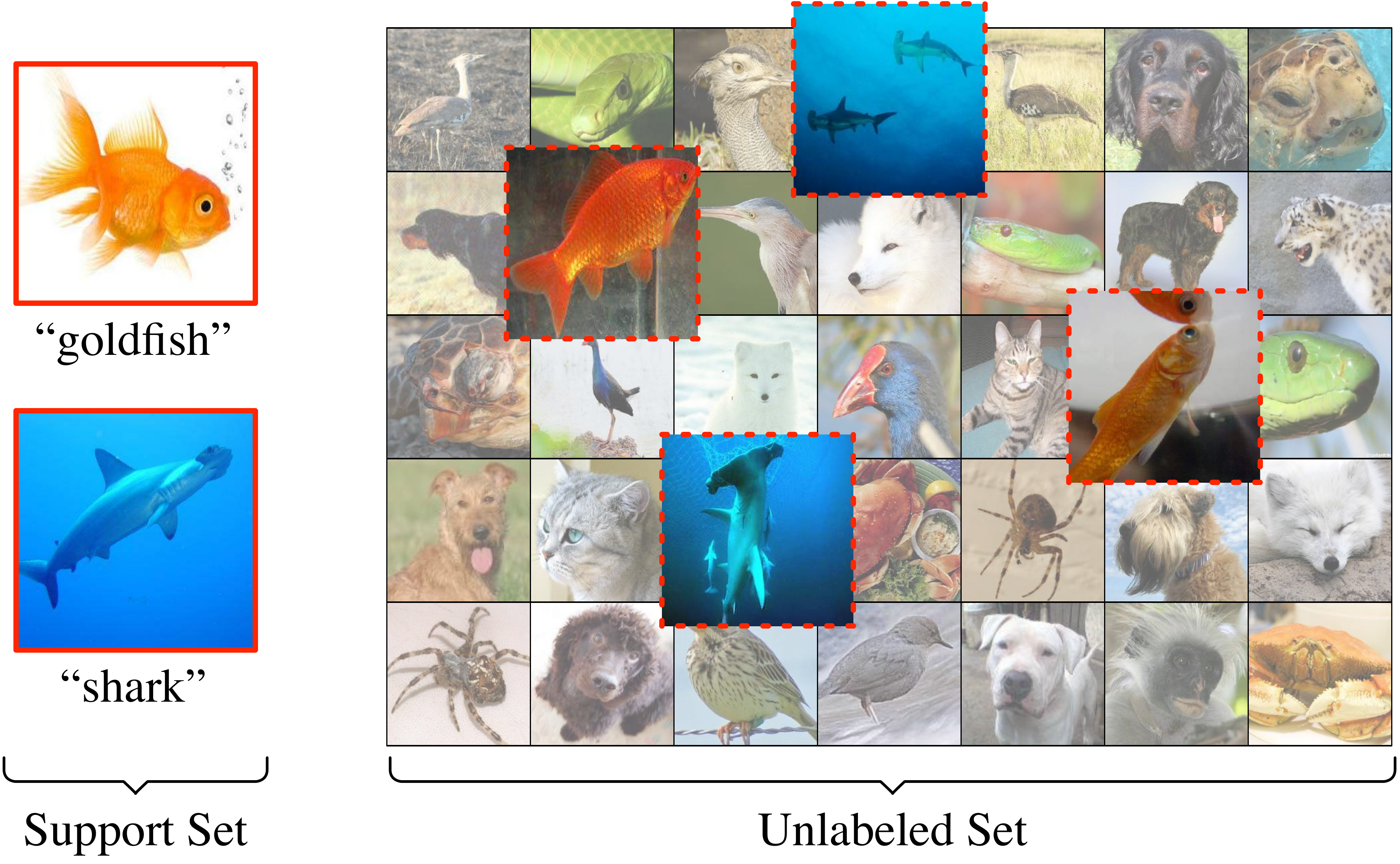}
    \caption{Consider a setup where the aim is to learn a classifier to distinguish between two
    previously unseen classes, goldfish and shark, given not only labeled examples of these two
    classes, but also a larger pool of unlabeled examples, some of which may belong to one of these
    two classes of interest. In this work we aim to move a step closer to this more natural learning
    framework by incorporating in our learning episodes unlabeled data from the classes we aim to
    learn representations for (shown with dashed red borders) as well as from {\it distractor}
    classes .}
    \label{fig:motivation}
    \vspace{-10pt}
\end{wrapfigure}

%% file: sections/background.tex
\section{Background}

We start by defining precisely the current paradigm for few-shot learning and the Prototypical
Network approach to this problem.

\subsection{Few-shot learning}

Recent progress on few-shot learning has been made possible by following an episodic paradigm.
Consider a situation where we have a large labeled dataset for a set of classes ${\cal C}_{\rm
train}$. However, after training on examples from ${\cal C}_{\rm train}$, our ultimate goal is to
produce classifiers for a disjoint set of new classes ${\cal C}_{\rm test}$, for which only a few
labeled examples will be available. The idea behind the episodic paradigm is to simulate the types
of few-shot problems that will be encountered at test, taking advantage of the large quantities of
available labeled data for classes ${\cal C}_{\rm train}$.


Specifically, models are trained on $K$-shot, $N$-way episodes constructed by first sampling a small subset of $N$ classes from ${\cal C}_{\rm train}$ and then generating:
1) a training (support) set ${\cal S}=\{(\bm{x}_1,y_1), (\bm{x}_2,y_2),
\dots, (\bm{x}_{N\times K},y_{N\times K})\}$ containing $K$ examples from each of the $N$ classes and 2) a test (query) set ${\cal Q}=\{(\bm{x}^*_1,y^*_1), (\bm{x}^*_2,y^*_2),
\dots, (\bm{x}^*_T,y^*_T)\}$ of different examples from the same $N$ classes. Each $\bm{x}_i \in \mathbb{R}^D$ is an input vector of
dimension $D$ and $y_i \in \{1, 2, \dots, N\}$ is a class label (similarly for $\bm{x}^*_i$ and $y^*_i$). Training on such episodes is done by feeding the support set ${\cal S}$ to the model and updating its parameters to minimize the loss of its predictions for the examples in the query set ${\cal Q}$.


One way to think of this approach is that our model  effectively trains to be a good learning algorithm. Indeed, much like a learning algorithm, the model must take in a set of labeled examples and produce a predictor that can be applied to new examples. Moreover, training directly encourages the classifier produced by the model to have good generalization on the new examples of the query set. Due to this analogy, training under this paradigm is often referred to as learning to learn or meta-learning. 

On the other hand, referring to the content of episodes as training and test sets and to the process of learning on these episodes as meta-learning or meta-training (as is sometimes done in the literature) can be confusing. So for the sake of clarity, we will refer to the content of episodes as support and query sets, and to the process of iterating over the training episodes simply as training.

\subsection{Prototypical Networks}

Prototypical Network~\citep{snell2017protonet} is a few-shot learning model that has the virtue  of being simple and yet obtaining state-of-the-art performance. At a high-level, it uses the support set ${\cal S}$ to extract a prototype vector from each class, and classifies the inputs in the query set based on their distance to the prototype of each class.

More precisely, Prototypical Networks learn an embedding function $h(\bm{x})$, parameterized as a neural network, that maps examples into a space where examples from the same class are close and those from different classes are far. All parameters of Prototypical Networks lie in the embedding function.

To compute the prototype $\bm{p}_c$ of each class $c$, a per-class average of the embedded examples is performed: 
\begin{align}
    \bm{p}_c = \frac{\sum_i h(\bm{x}_i) z_{i,c}}{\sum_i z_{i,c}},~~{\rm where }~~z_{i,c} = \mathbbm{1}[y_i = c].~\label{eq:prototypes}
\end{align}
These prototypes define a predictor for the class of any new (query) example $\bm{x}^*$, which assigns a probability over any class $c$ based on the distances between $\bm{x}^*$ and each prototype, as follows:
\begin{align}
    p(c|\bm{x}^*,\{\bm{p}_c\}) = \frac{\exp(-||h(\bm{x}^*) - \bm{p}_c||^2_2)}{\sum_{c'}\exp(-||h(\bm{x}^*) - \bm{p}_{c'}||^2_2)}~.\label{eq:classprobs}
\end{align}
The loss function used to update Prototypical Networks for a given training episode is then simply the average negative log-probability of the correct class assignments, for all query examples:
\begin{align}
    -\frac{1}{T}\sum_{i} \log p(y^*_i|\bm{x}_i^*,\{\bm{p}_c\})~.\label{eq:loss}
\end{align}
Training proceeds by minimizing the average loss, iterating over training episodes and performing a gradient descent update for each. 

Generalization performance is measured on test set episodes, which contain images from classes in ${\cal C}_{\rm test}$ instead of ${\cal C}_{\rm train}$. For each test episode, we use the predictor produced by the Prototypical Network for the provided support set ${\cal S}$ to classify each of query input $\bm{x}^*$ into the most likely class $\hat{y} = \argmax_c p(c|\bm{x}^*,\{\bm{p}_c\})$.

%% file: sections/model.tex
\section{Semi-Supervised Few-Shot Learning}

We now define the semi-supervised setting considered in this work for few-shot learning.

The training set is denoted as a tuple of labeled and unlabeled examples: $(\mathcal{S},
\mathcal{R})$. The labeled portion is the usual support set $\mathcal{S}$ of the few-shot learning
literature, containing a list of tuples of inputs and targets.  In addition to classic few-shot
learning, we introduce an unlabeled set $\mathcal{R}$ containing only inputs:
$\mathcal{R}=\{\tilde{\bm{x}}_1, \tilde{\bm{x}}_2, \dots, \tilde{\bm{x}}_M\}$.  As in the purely
supervised setting, our models are trained to perform well when predicting the labels for the
examples in the episode's query set $\mathcal{Q}$.  Figure~\ref{fig:episode_setup} shows a
visualization of training and test episodes.

\input figures/episode_figure

\subsection{Semi-Supervised Prototypical Networks}

In their original formulation, Prototypical Networks do not specify a way to leverage the unlabeled
set $\mathcal{R}$. In  what follows, we now propose various extensions that start from the basic
definition of prototypes $\bm{p}_c$ and provide a procedure for producing refined prototypes
$\tilde{\bm{p}}_c$ using the unlabeled examples in $\mathcal{R}$.

\input figures/refinement_figure

After the refined prototypes are obtained, each of these models is trained with the same loss
function for ordinary Prototypical Networks of Equation~\ref{eq:loss}, but replacing $\bm{p}_c$
with $\tilde{\bm{p}}_c$. That is, each query example is classified into one of the $N$ classes based
on the proximity of its embedded position with the corresponding {\it refined} prototypes, and the
average negative log-probability of the correct classification is used for training.

\subsubsection{Prototypical Networks with Soft $k$-Means}
We first consider a simple way of leveraging unlabeled examples for refining prototypes, by taking
inspiration from semi-supervised clustering. Viewing each prototype as a cluster center, the
refinement process could attempt to adjust the cluster locations to better fit the examples in both
the support and unlabeled sets. Under this view, cluster assignments of the labeled examples in the
support set are considered known and fixed to each example's label.  The refinement process must
instead estimate the cluster assignments of the unlabeled examples and adjust the cluster locations
(the prototypes) accordingly.

One natural choice would be to borrow from the inference performed by soft $k$-means. We prefer this
version of $k$-means over hard assignments since hard assignments would make the inference 
non-differentiable.  We start with the regular Prototypical Network's prototypes $\bm{p}_{c}$ (as
specified in Equation~\ref{eq:prototypes}) as the cluster locations. Then, the unlabeled examples
get a partial assignment ($\tilde{z}_{j,c}$) to each cluster based on their Euclidean distance to
the cluster locations. Finally, refined  prototypes are obtained by incorporating these unlabeled
examples.

This process can be summarized as follows:
\begin{align}
    \tilde{\bm{p}}_c = \frac{\sum_i h(\bm{x}_i) z_{i,c} + \sum_j h(\tilde{\bm{x}}_j) \tilde{z}_{j,c}}
    {\sum_i z_{i,c} + \sum_j \tilde{z}_{j,c}}, ~~{\rm where }~~
    \tilde{z}_{j,c} = \frac{\exp \left(-||h(\tilde{\bm{x}}_j) - \bm{p}_c||^2_2 \right)}
    {\sum_{c'} \exp \left(-||h(\tilde{\bm{x}}_j) - \bm{p}_{c'}||^2_2 \right) } \label{eq:softassign}
\end{align}
Predictions of each query input's class is then modeled as in Equation~\ref{eq:classprobs}, but 
using the refined prototypes $\tilde{\bm{p}}_c$.

We could perform several iterations of refinement, as is usual in $k$-means. However, we have
experimented with various number of iterations and found results to not improve beyond a single
refinement step.

\subsubsection{Prototypical Networks with Soft $k$-Means with a Distractor Cluster}
The soft $k$-means approach described above implicitly assumes that each unlabeled example belongs
to either one of the $N$ classes in the episode. However, it would be much more general to not make
that assumption and have a model robust to the existence of examples from other classes, which we
refer to as distractor classes. For example, such a situation would arise if we wanted to distinguish
between pictures of unicycles and scooters, and decided to add an unlabeled set  by downloading
images from the web. It then would not be realistic to assume that all these images are of unicycles
or scooters. Even with a focused search, some may be from similar classes, such as bicycle.

Since soft $k$-means distributes its soft assignments across all classes, distractor items could be
harmful and interfere with the refinement process, as prototypes would be adjusted to also partially
account for these distractors. A simple way to address this is to add an additional cluster whose
purpose is to capture the distractors, thus preventing them from polluting the clusters of the
classes of interest:
\begin{equation}
    \bm{p}_c =
    \begin{cases}
        \frac{\sum_i h(\bm{x}_i) z_{i,c}}{\sum_i z_{i,c}} & \text{\ \ for\ \ } c = 1...N\\
        \bm{0} & \text{\ \ for\ \ } c = N+1
    \end{cases}
\end{equation}
Here we take the simplifying assumption that the distractor cluster has a prototype centered at the
origin.  We also consider introducing length-scales $r_c$ to represent variations in the 
within-cluster distances, specifically for the distractor cluster:
\begin{align}
    \tilde{z}_{j,c} = \frac{\exp \left(-\frac{1}{r_c^2}||\tilde{\bm{x}}_j - \bm{p}_c||^2_2 -A(r_c) \right)}
    {\sum_{c'} \exp \left(-\frac{1}{r_c^2}||\tilde{\bm{x}}_j - \bm{p}_{c'}||^2_2 -A(r_{c'}) \right) }, ~~{\rm where}~~
    A(r) = \frac{1}{2}\log(2\pi) + \log(r)
\end{align}
For simplicity, we set $r_{1\dots N}$ to 1 in our experiments, and only learn the
length-scale of the distractor cluster $r_{N+1}$.

\subsubsection{Prototypical Networks with Soft $k$-Means and Masking}
Modeling distractor unlabeled examples with a single  cluster is likely too simplistic. Indeed, it
is inconsistent with our assumption that each cluster corresponds to one class, since distractor
examples may very well cover more than a single natural object category.  Continuing with our
unicycles and bicycles example, our web search for unlabeled images could accidentally include not
only bicycles, but other related objects such as tricycles or cars.   This was also reflected in our
experiments, where we constructed the episode generating process so that it would sample distractor
examples from multiple classes.

To address this problem, we propose an improved variant: instead of capturing distractors with a
high-variance catch-all cluster, we model distractors as examples that are not within some area of
any of the legitimate class prototypes. This is done by incorporating a soft-masking mechanism on
the contribution of unlabeled examples. At a high level, we want unlabeled examples that are closer
to a prototype to be masked less than those that are farther.

More specifically, we modify the soft $k$-means refinement as follows. We start by computing 
normalized distances $\tilde{d}_{j,c}$ between examples $\tilde{\bm{x}}_j$ and prototypes $\bm{p}_c$: 
\begin{align}
    \tilde{d}_{j,c}=\frac{d_{j,c}}
    {\frac{1}{M} \sum_j d_{j,c}},&~~{\rm where }~~ d_{j,c} = ||h(\tilde{\bm{x}}_j) - \bm{p}_c||^2_2
\end{align}
Then, soft thresholds $\beta_c$ and slopes $\gamma_c$ are predicted for each prototype, by feeding 
to a small neural network various statistics of the normalized distances for the prototype:
\begin{align}
    \left[\beta_c, \gamma_c \right] &= {\rm MLP} \left(
    \left[\min_j(\tilde{d}_{j,c}), \max_j(\tilde{d}_{j,c}), 
    \var_j(\tilde{d}_{j,c}), \skewness_j(\tilde{d}_{j,c}), 
    \kurt_j(\tilde{d}_{j,c}) \right] \right) \label{eq:masks}
\end{align}
This allows each threshold to use information on the amount of intra-cluster variation to determine 
how aggressively it should cut out unlabeled examples.

Then, soft masks $m_{j,c}$ for the contribution of each example to each prototype are computed, by 
comparing to the threshold the normalized distances, as follows:
\begin{align}
\tilde{\bm{p}}_c = \frac{\sum_i h(\bm{x}_i) z_{i,c} + \sum_j h(\tilde{\bm{x}}_j) \tilde{z}_{j,c} m_{j,c}}
                             {\sum_i z_{i,c} + \sum_j \tilde{z}_{j,c} m_{j,c}},
    &~~{\rm where }~~ m_{j,c} = \sigma \left(-\gamma_c \left(
    \tilde{d}_{j,c}-\beta_c \right) \right)
\end{align}
where $\sigma(\cdot)$ is the sigmoid function. 

When training with this refinement process, the model can now use its MLP in Equation~\ref{eq:masks}
to learn to include or ignore entirely certain unlabeled examples. The use of soft masks makes this
process entirely differentiable\footnote{We stop gradients from passing through the computation of
the statistics in Equation~\ref{eq:masks}, to avoid potential numerical instabilities.}.
Finally, much like for regular soft $k$-means (with or without a distractor cluster), while we could
recursively repeat the refinement for multiple steps, we found a single step to perform well enough.

%% file: figures/episode_figure.tex

\begin{figure}[ht]
    \centering
    \includegraphics[width=0.77\textwidth]{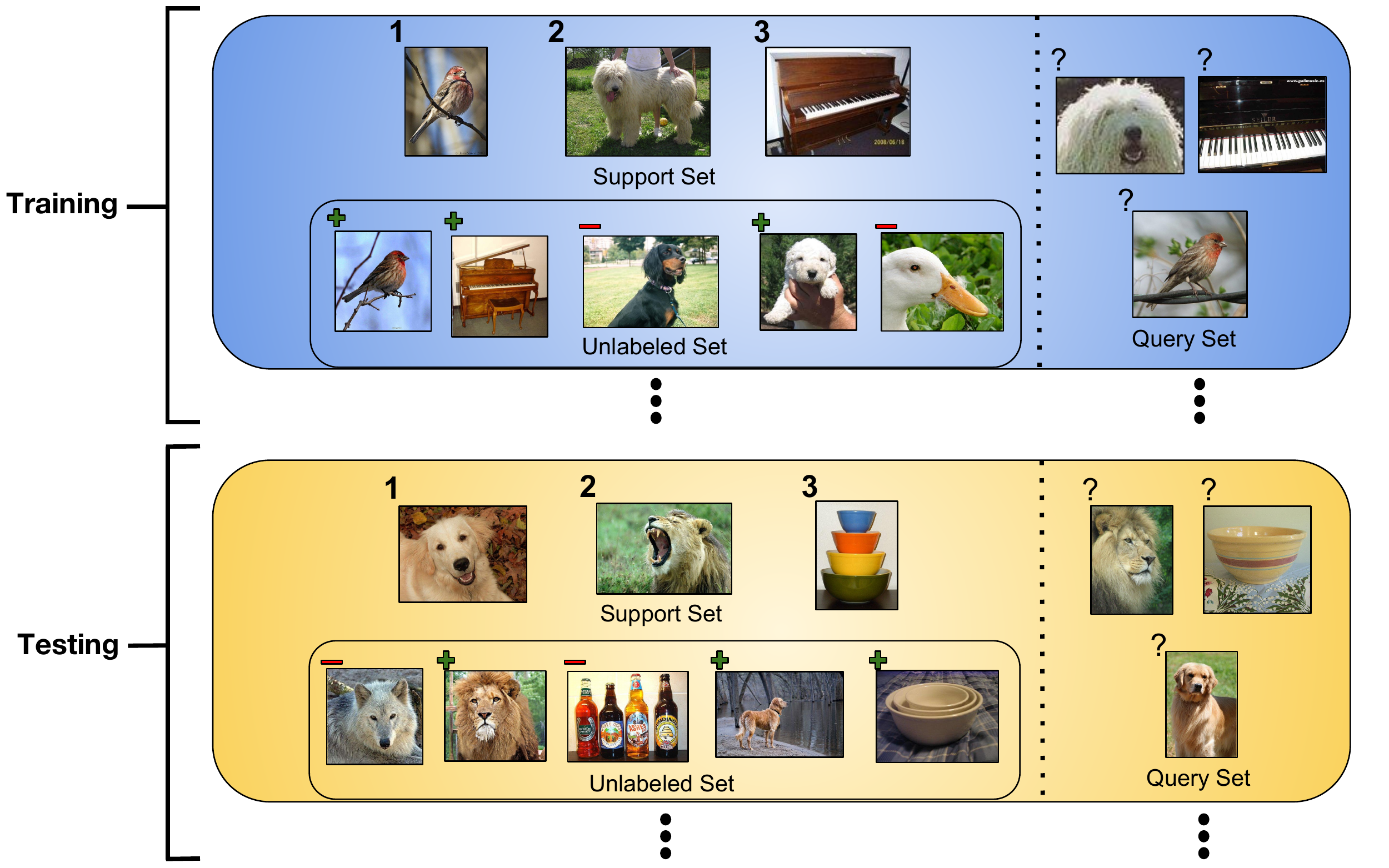}
    \caption{Example of the semi-supervised few-shot learning setup. Training involves iterating through training episodes, consisting of a support set $\mathcal{S}$, an unlabeled set $\mathcal{R}$, and a query set $\mathcal{Q}$. The goal is to use the labeled items (shown with their numeric class label) in $\mathcal{S}$ and the unlabeled items in $\mathcal{R}$ within each episode to generalize to good performance on the corresponding query set. The unlabeled items in $\mathcal{R}$ may either be pertinent to the classes we are considering (shown above with green plus signs) or they may be \emph{distractor} items which belong to a class that is not relevant to the current episode (shown with red minus signs). However note that the model does not actually have ground truth information as to whether each unlabeled example is a distractor or not; the plus/minus signs are shown only for illustrative purposes. At test time, we are given new episodes consisting of novel classes not seen during training that we use to evaluate the meta-learning method.}
    \label{fig:episode_setup}
\end{figure}

%% file: figures/refinement_figure.tex
\begin{wrapfigure}{r}{0.65\textwidth}
    \centering
    \vspace{-0.1in}
    \includegraphics[width=0.6\textwidth]{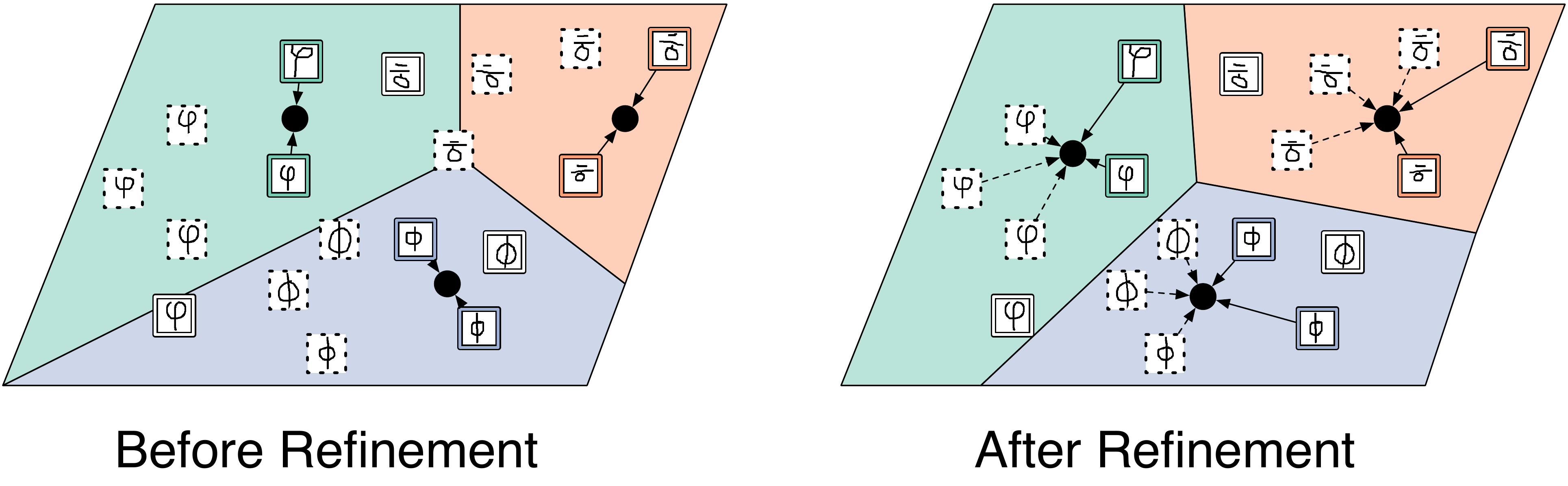}
    \caption{Left: The prototypes are initialized based on the mean location of the examples of the corresponding class, as in ordinary Prototypical Networks. Support, unlabeled, and query examples have solid, dashed, and white colored borders respectively. Right: The refined prototypes obtained by incorporating the unlabeled examples, which classifies all query examples correctly.}
    \label{fig:refinement}
    \vspace{-15pt}
\end{wrapfigure}

%% file: sections/related.tex
\section{Related Work}

We summarize here the most relevant work from the literature on few-shot learning, semi-supervised learning and clustering.

The best performing methods for few-shot learning use the episodic training framework prescribed by
meta-learning. The approach within which our work falls is that of metric learning methods. Previous
work in metric-learning for few-shot-classification includes Deep Siamese
Networks~\citep{koch2015siamese}, Matching Networks~\citep{vinyals2016matchingnet}, and Prototypical
Networks~\citep{snell2017protonet}, which is the model we extend to the semi-supervised setting in
our work. The general idea here is to learn an embedding function that embeds examples belonging to
the same class close together while keeping embeddings from separate classes far apart. Distances
between embeddings of items from the support set and query set are then used as a notion of
similarity to do classification. Lastly, closely related to our work with regard to extending the
few-shot learning setting,~\cite{bachman2017active-learning} employ Matching Networks in an active
learning framework where the model has a choice of which unlabeled item to add to the support set
over a certain number of time steps before classifying the query set. Unlike our setting, their
meta-learning agent can acquire ground-truth labels from the unlabeled set, and they do not use
distractor examples.

Other meta-learning approaches to few-shot learning include learning how to use the support set to
update a learner model so as to generalize to the query set. Recent work has involved learning
either the weight initialization and/or update step that is used by a learner neural
network~\citep{ravi2017oneshot,FinnC2017}. Another approach is to train a generic neural
architecture such as a memory-augmented recurrent network~\citep{Santoro2016} or a temporal
convolutional network~\citep{MishraN2017} to sequentially process the support set and perform
accurate predictions of the labels of the query set examples. These other methods are also
competitive for few-shot learning, but we chose to extend Prototypical Networks in this work for its
simplicity and efficiency.

As for the literature on semi-supervised learning, while it is quite vast~\citep{zhu2005semi,
ChapelleO2010}, the most relevant category to our work is related to 
self-training~\citep{yarowsky1995unsupervised, rosenberg2005semi}. Here, a classifier is first trained on
the initial training set. The classifier is then used to classify unlabeled items, and the most
confidently predicted unlabeled items are added to the training set with the prediction of the
classifier as the assumed label.  This is similar to our soft $k$-Means extension to Prototypical
Networks. Indeed, since the soft assignments (Equation~\ref{eq:softassign}) match the regular
Prototypical Network's classifier output for new inputs (Equation~\ref{eq:classprobs}), then the
refinement can be thought of re-feeding to a Prototypical Network a new support set  augmented with
(soft) self-labels from the unlabeled set.

Our algorithm is also related to transductive 
learning~\citep{vapnik1998statistical,Joachims1999TSVM,Fu2015TransductiveZSL}, where the base
classifier gets refined by seeing the unlabeled examples. In practice, one could use our method in a
transductive setting where the unlabeled set is the same as the query set; 
however, here to avoid our model memorizing labels of the unlabeled set during the meta-learning 
procedure, we split out a separate unlabeled set that is different from the query set.

In addition to the original $k$-Means method~\citep{lloyd1982least}, the most related work to our
setup involving clustering algorithms considers applying $k$-Means in the presence of
outliers~\citep{hautamaki2005improving, chawla2013k, gupta2017local}. The goal here is to correctly
discover and ignore the outliers so that they do not wrongly shift the cluster locations to form a
bad partition of the true data. This objective is also important in our setup as not ignoring
outliers (or distractors) will wrongly shift the prototypes and negatively influence classification
performance.

Our contribution to the semi-supervised learning and clustering literature is to go beyond the
classical setting of training and evaluating within a single dataset, and consider the setting where
we must learn to transfer from a set of training classes ${\cal C}_{\rm train}$ to a new set of test
classes ${\cal C}_{\rm test}$.

%% file: sections/experiments.tex
\section{Experiments}

\subsection{Datasets}

We evaluate the performance of our model on three datasets: two benchmark few-shot classification
datasets and a novel large-scale dataset that we hope will be useful for future few-shot learning
work.

\textbf{Omniglot} \citep{lake2011oneshot} is a dataset of 1,623 handwritten characters from 50
alphabets. Each character was drawn by 20 human subjects. We follow the few-shot setting proposed by
\citet{vinyals2016matchingnet}, in which the images are resized to $28 \times 28$ pixels and
rotations in multiples of 90$^\circ$ are applied, yielding 6,492 classes in total. These are split
into 4,112 training classes, 688 validation classes, and 1,692 testing classes.

\textbf{\textit{mini}ImageNet} \citep{vinyals2016matchingnet} is a modified version of the ILSVRC-12
dataset \citep{russakovsky2015imagenet}, in which 600 images for each of 100 classes were randomly
chosen to be part of the dataset. We rely on the class split used by \citet{ravi2017oneshot}. These
splits use 64 classes for training, 16 for validation, and 20 for test. All images are of size 84
$\times$ 84 pixels.

\textbf{\textit{tiered}ImageNet} is our proposed dataset for few-shot classification. Like
\textit{mini}Imagenet, it is a subset of ILSVRC-12. However, \textit{tiered}ImageNet represents a
larger subset of ILSVRC-12 (608 classes rather than 100 for \textit{mini}ImageNet). Analogous to
Omniglot, in which characters are grouped into alphabets, \textit{tiered}ImageNet groups classes
into broader categories corresponding to higher-level nodes in the ImageNet \citep{deng2009imagenet}
hierarchy. There are 34 categories in total, with each category containing between 10 and 30
classes. These are split into 20 training, 6 validation and 8 testing categories (details of the
dataset can be found in the supplementary material). This ensures that all of the training classes
are sufficiently distinct from the testing classes, unlike \textit{mini}ImageNet and other
alternatives such as \textit{rand}ImageNet proposed by  \citet{vinyals2016matchingnet}. For example,
``pipe organ'' is a training class and ``electric guitar'' is a test class in the
\citet{ravi2017oneshot} split of  \textit{mini}Imagenet, even though they are both musical
instruments. This scenario would not occur in \textit{tiered}ImageNet since ``musical instrument''
is a high-level category and as such is not split between training and test classes. This represents
a more realistic few-shot learning scenario since in general we cannot assume that test classes will
be similar to those seen in training. Additionally, the tiered structure of \textit{tiered}ImageNet
may be useful for few-shot learning approaches that can take advantage of hierarchical relationships
between classes. We leave such interesting extensions for future work.

\subsection{Adapting the Datasets for Semi-Supervised Learning}
For each dataset, we first create an additional split to separate the images of each class into
disjoint labeled and unlabeled sets. For Omniglot and {\it tiered}ImageNet we sample 10\% of the
images of each class to form the labeled split. The remaining 90\% can only be used in the unlabeled
portion of episodes. For {\it mini}ImageNet we use 40\% of the data for the labeled split
and the remaining 60\% for the unlabeled, since we noticed that 10\% was too small to achieve
reasonable performance and avoid overfitting. We report the average classification scores over 10
random splits of labeled and unlabeled portions of the training set, with uncertainty computed in
standard error (standard deviation divided by the square root of the total number of splits).

We would like to emphasize that due to this labeled/unlabeled split, we are using strictly less
label information than in the previously-published work on these datasets. Because of this, we do
not expect our results to match the published numbers, which should instead be interpreted as an
upper-bound for the performance of the semi-supervised models defined in this work.

Episode construction then is performed as follows. For a given dataset, we create a training episode
by first sampling $N$ classes uniformly at random from the set of training classes ${\cal C}_{\rm
train}$. We then sample $K$ images from the labeled split of each of these classes to form the
support set, and $M$ images from the unlabeled split of each of these classes to form the unlabeled
set. Optionally, when including distractors, we additionally sample $H$ other classes from the set
of training classes and $M$ images from the unlabeled split of each to act as the distractors. These
distractor images are added to the unlabeled set along with the unlabeled images of the $N$ classes
of interest (for a total of $MN + MH$ unlabeled images). The query portion of the episode is comprised of a
fixed number of images from the labeled split of each of the $N$ chosen classes. Test episodes are
created analogously, but with the $N$ classes (and optionally the $H$ distractor classes) sampled
from ${\cal C}_{\rm test}$. In the experiments reported here we used
$H=N=5$, i.e.\ 5 classes for both the labeled classes and the distractor classes.  We used $M=5$ for training and $M=20$ for testing in most cases, thus
measuring the ability of the models to generalize to a larger unlabeled set size. 
Details of the dataset splits, including the specific classes assigned to train/validation/test sets, can be found in Appendices A and B.

In each dataset we compare our three semi-supervised models with two baselines. The first baseline,
referred to as ``Supervised'' in our tables, is an ordinary Prototypical Network that is trained in
a purely supervised way on the labeled split of each dataset. The second baseline, referred to as
``Semi-Supervised Inference'', uses the embedding function learned by this supervised Prototypical
Network, but performs semi-supervised refinement of the prototypes at test time using a step of
Soft $k$-Means refinement. This is to be contrasted with our semi-supervised models that perform
this refinement both at training time and at test time, therefore learning a different embedding
function. We evaluate each model in two settings: one where all unlabeled examples belong to the
classes of interest, and a more challenging one that includes distractors. Details of the model
hyperparameters can be found in Appendix~\ref{sec:hyperparam} and our online repository.\footnote{
Code available at
\url{https://github.com/renmengye/few-shot-ssl-public}}

\input{sections/results}

%% file: sections/results.tex
\subsection{Results}
\input{sections/results_omniglot}

\input{sections/results_miniImagenet}

\input{sections/results_tieredImagenet}

Results for Omniglot, {\it mini}ImageNet and {\it tiered}ImageNet are given in
Tables~\ref{tab:omniglot},~\ref{tab:miniImageNet} and~\ref{tab:tieredImageNet}, respectively, while
Figure~\ref{fig:tnet_num_unlabel} shows the performance of our models on {\it tiered}ImageNet (our
largest dataset) using different values for $M$ (number of items in the unlabeled set per class).
Additional results comparing the ProtoNet model to various baselines on these datasets, and analysis of the performance of the Masked Soft $k$-Means model can be found in Appendix C.

Across all three benchmarks, at least one of our proposed models outperforms the baselines,
demonstrating the effectiveness of our semi-supervised meta-learning procedure. 
In the non-distractor settings, all three proposed models outperform the baselines in almost all the experiments, without a clear winner between the three models across the datasets and shot numbers.
In the scenario where training and testing includes distractors, Masked Soft $k$-Means shows the most robust performance across all
three datasets, attaining the best results in each case but one. In fact this model reaches performance that is close to the upper bound based on the results without distractors.


From Figure~\ref{fig:tnet_num_unlabel}, we observe clear improvements in test accuracy when the
number of items in the unlabeled set per class grows from 0 to 25. These models were trained with $M=5$ and thus are showing an
ability to extrapolate in generalization. This confirms that, through meta-training, the models
learn to acquire a better representation that is improved by semi-supervised
refinement.


%% file: sections/results_omniglot.tex
\begin{table}[t]
    \centering
    \begin{tabular}{l|c|c}
    Models                             & Acc.                  & Acc.  w/ D              \\
    \hline\hline
    Supervised                         & 94.62 $\pm$ 0.09      & 94.62 $\pm$ 0.09        \\
    \hline
    Semi-Supervised Inference          & 97.45 $\pm$ 0.05      & 95.08 $\pm$ 0.09        \\
    \hline\hline
    Soft $k$-Means                     & 97.25 $\pm$ 0.10      & 95.01 $\pm$ 0.09        \\
    \hline
    Soft $k$-Means+Cluster             & \tb{97.68 $\pm$ 0.07} & 97.17 $\pm$ 0.04        \\
    \hline
    Masked Soft $k$-Means              & 97.52 $\pm$ 0.07      & \tb{97.30 $\pm$ 0.08}   \\
    \end{tabular}
    \caption{Omniglot 1-shot classification results. In this table as well as those below ``w/ D'' denotes ``with distractors'', where the
    unlabeled images contain irrelevant classes.}
    \label{tab:omniglot}
\end{table}

%% file: sections/results_miniImagenet.tex
\begin{table}[t]
    \centering
    \resizebox{\textwidth}{!}{
    \begin{tabular}{l|c|c|c|c}
    Models                       & 1-shot Acc.         & 5-shot Acc.         & 1-shot Acc w/ D     & 5-shot Acc. w/ D \\
    \hline\hline
    Supervised                   & 43.61 $\pm$ 0.27    & 59.08 $\pm$ 0.22   & 43.61 $\pm$ 0.27    & 59.08 $\pm$ 0.22 \\
    \hline
    Semi-Supervised Inference    & 48.98 $\pm$ 0.34    & 63.77 $\pm$ 0.20    & 47.42 $\pm$ 0.33    & 62.62 $\pm$ 0.24 \\
    \hline\hline
    Soft $k$-Means               &\tb{50.09 $\pm$ 0.45}&\tb{64.59 $\pm$ 0.28}&\tb{48.70 $\pm$ 0.32}&\tb{63.55 $\pm$ 0.28}\\
    \hline
    Soft $k$-Means+Cluster       & 49.03 $\pm$ 0.24    & 63.08 $\pm$ 0.18    &\tb{48.86 $\pm$ 0.32}& 61.27 $\pm$ 0.24 \\
    \hline
    Masked Soft $k$-Means        &\tb{50.41 $\pm$ 0.31}&\tb{64.39 $\pm$ 0.24}&\tb{49.04 $\pm$ 0.31}& 62.96 $\pm$ 0.14 \\
    \end{tabular}
    }
    \caption{{\it mini}ImageNet 1/5-shot classification results.}
    \label{tab:miniImageNet}
\end{table}

%% file: sections/results_tieredImagenet.tex
\begin{table}[t]
    \centering
    \resizebox{\textwidth}{!}{
    \begin{tabular}{l|c|c|c|c}
    Models                    & 1-shot Acc.         & 5-shot Acc.         & 1-shot Acc. w/ D    & 5-shot Acc. w/ D     \\
    \hline\hline
    Supervised                & 46.52 $\pm$ 0.52    & 66.15 $\pm$ 0.22    & 46.52 $\pm$ 0.52    & 66.15 $\pm$ 0.22     \\
    \hline
    Semi-Supervised Inference & 50.74 $\pm$ 0.75    & 69.37 $\pm$ 0.26    & 48.67 $\pm$ 0.60    & 67.46 $\pm$ 0.24      \\
    \hline\hline
    Soft $k$-Means            & 51.52 $\pm$ 0.36    &\tb{70.25 $\pm$ 0.31}& 49.88 $\pm$ 0.52    & 68.32 $\pm$ 0.22     \\
    \hline
    Soft $k$-Means+Cluster    &\tb{51.85 $\pm$ 0.25}& 69.42 $\pm$ 0.17    &\tb{51.36 $\pm$ 0.31}& 67.56 $\pm$ 0.10     \\
    \hline
    Masked Soft $k$-Means     &\tb{52.39 $\pm$ 0.44}&\tb{69.88 $\pm$ 0.20}&\tb{51.38 $\pm$ 0.38}&\tb{69.08 $\pm$ 0.25} \\
    \end{tabular}
    }
    \caption{\textit{tiered}ImageNet 1/5-shot classification results.}
    \label{tab:tieredImageNet}
\end{table}
\begin{figure}
    \centering
    \includegraphics[width=\textwidth]{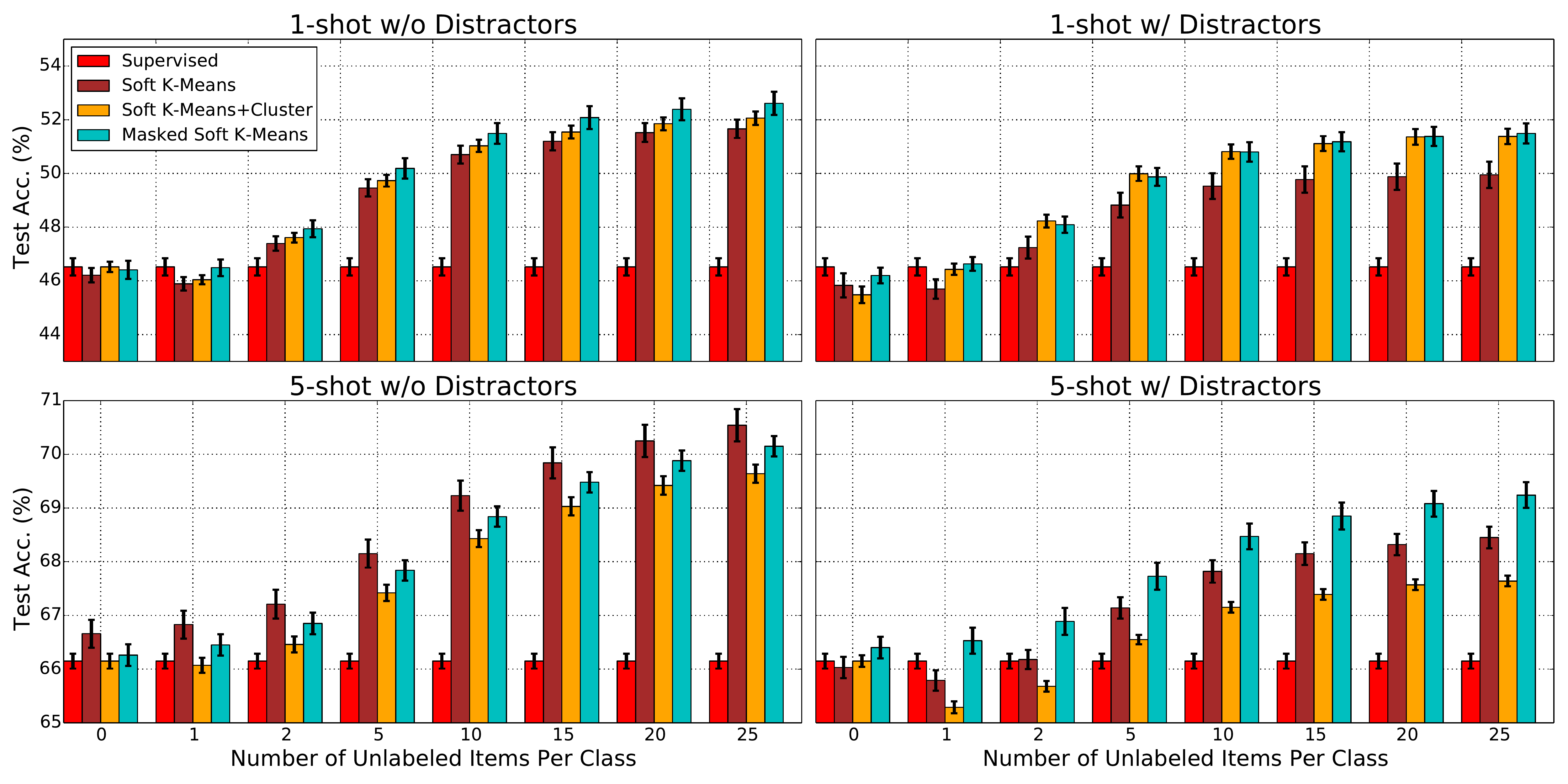}
    \caption{Model Performance on {\it tiered}ImageNet with different numbers of unlabeled items during test time.}
    \label{fig:tnet_num_unlabel}
\end{figure}

%% file: sections/conclusion.tex
\section{Conclusion}

In this work, we propose a novel semi-supervised few-shot learning paradigm, where an unlabeled set
is added to each episode. We also extend the setup to more realistic situations where the unlabeled
set has novel classes distinct from the labeled classes. To address the problem that current few-shot
classification datasets are too small for a labeled vs.\ unlabeled split and also lack
hierarchical levels of labels, we introduce a new dataset, \textit{tiered}ImageNet. We propose
several novel extensions of Prototypical Networks, and they show consistent improvements under 
semi-supervised settings compared to our baselines. As future work, we are working on incorporating 
fast weights \citep{ba2016fw,FinnC2017} into our framework so that examples can have different
embedding representations given the contents in the episode.

%% file: sections/acknowledgement.tex
\paragraph{Acknowledgement}
Supported by grants from NSERC, Samsung, and the Intelligence Advanced Research Projects Activity (IARPA) via Department of
Interior/Interior Business Center (DoI/IBC) contract number D16PC00003. The U.S. Government is
authorized to reproduce and distribute reprints for Governmental purposes notwithstanding any
copyright annotation thereon. Disclaimer: The views and conclusions contained herein are those of
the authors and should not be interpreted as necessarily representing the official policies or
endorsements, either expressed or implied, of IARPA, DoI/IBC, or the U.S. Government.

%% file: sections/supplementary.tex
\section{Omniglot Dataset Details}
We used the following split details for experiments on Omniglot dataset. This is the same train/test split as \citep{vinyals2016matchingnet}, but we created our own validation split for selecting hyper-parameters. Models are trained on the train split only.

\textbf{Train Alphabets:}
Alphabet\_of\_the\_Magi,
Angelic,
Anglo-Saxon\_Futhorc,
Arcadian,
Asomtavruli\_(Georgian),
Atemayar\_Qelisayer,
Atlantean,
Aurek-Besh,
Avesta,
Balinese,
Blackfoot\_(Canadian\_Aboriginal\_Syllabics),
Braille,
Burmese\_(Myanmar),
Cyrillic,
Futurama,
Ge\_ez,
Glagolitic,
Grantha,
Greek,
Gujarati,
Gurmukhi (character 01-41),
Inuktitut\_(Canadian\_Aboriginal\_Syllabics),
Japanese\_(hiragana),
Japanese\_(katakana),
Korean,
Latin,
Malay\_(Jawi\_-\_Arabic),
N\_Ko,
Ojibwe\_(Canadian\_Aboriginal\_Syllabics),
Sanskrit,
Syriac\_(Estrangelo),
Tagalog,
Tifinagh

\textbf{Validation Alphabets:}
Armenian, 
Bengali, 
Early\_Aramaic, 
Hebrew, 
Mkhedruli\_(Geogian) 

\textbf{Test Alphabets:}
Gurmukhi (character 42-45),
Kannada,
Keble,
Malayalam,
Manipuri,
Mongolian,
Old\_Church\_Slavonic\_(Cyrillic),
Oriya,
Sylheti,
Syriac\_(Serto),
Tengwar,
Tibetan,
ULOG

\section{\textit{tiered}Imagenet Dataset Details}

\begin{sidewaysfigure}[ht]
    \includegraphics[width=\textwidth]{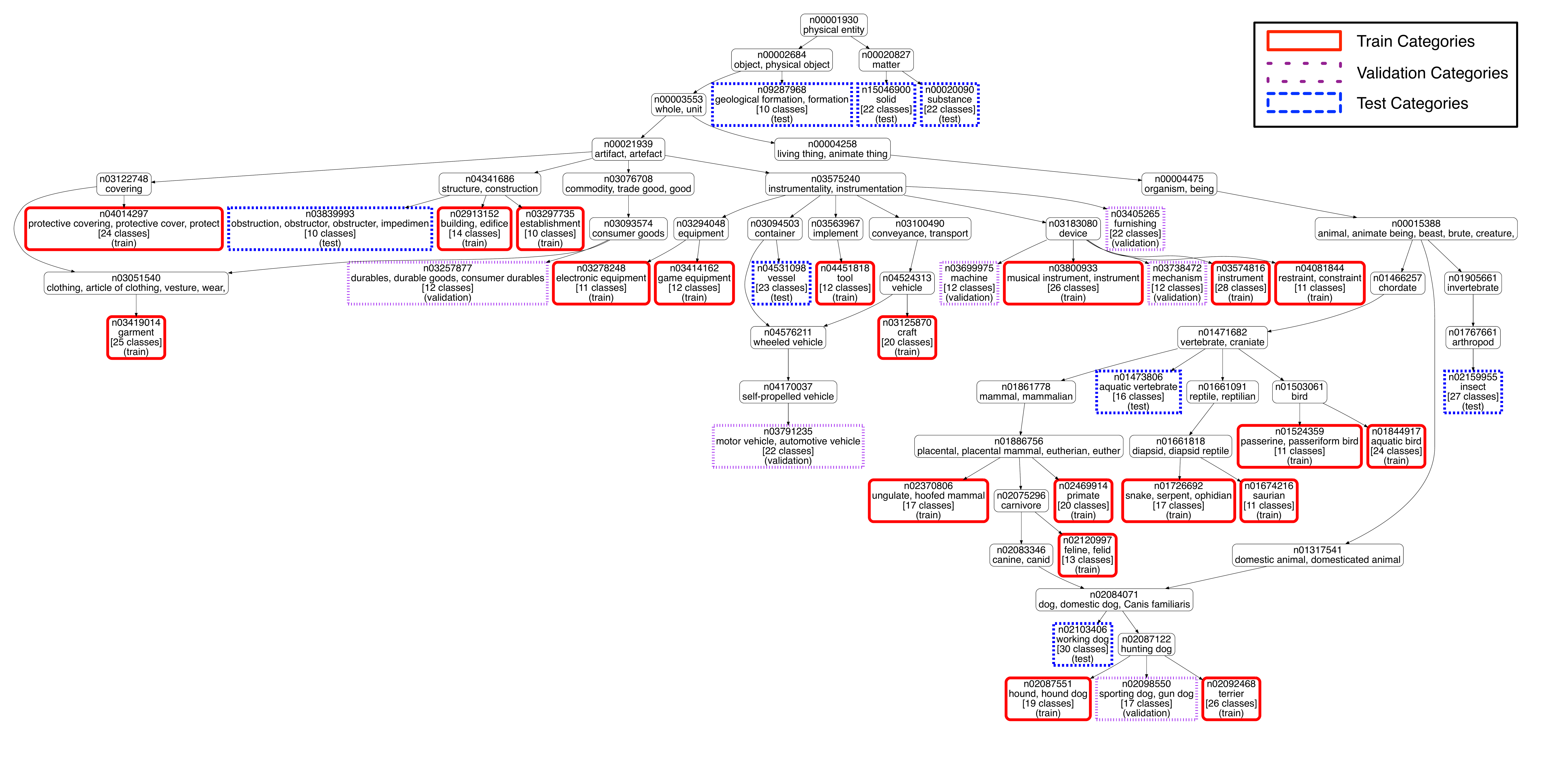}
    \caption{Hierarchy of \textit{tiered}Imagenet categories. Training categories are highlighted in red and test categories in blue. Each category indicates the number of associated classes from ILSVRC-12. Best viewed zoomed-in on electronic version.}
    \label{fig:tiered_hierarchy}
\end{sidewaysfigure}

Each high-level category in \textit{tiered}ImageNet contains between 10 and 30 ILSVRC-12 classes (17.8 on average). In the ImageNet hierarchy, some classes have multiple parent nodes. Therefore, classes belonging to more than one category were removed from the dataset to ensure separation between training and test categories. Test
categories were chosen to reflect various levels of separation between training
and test classes. Some test categories (such as ``working dog'') are fairly similar to training categories, whereas others (such as ``geological formation'')
are quite different. The list of categories is shown below and statistics of the dataset can be found in Table~\ref{tab:tiered_stats}. A visualization of the categories according to the ImageNet hierarchy is shown in Figure~\ref{fig:tiered_hierarchy}. The full list of classes per category will
also be made public, however for the sake of brevity we do not include it here.

\begin{table}[ht]
    \center
    \caption{Statistics of the \textit{tiered}ImageNet dataset.}
    \label{tab:tiered_stats}
    \begin{tabular}{l|cccc}
                   & Train   & Val & Test    & Total \\ \hline
        Categories & 20      & 6   & 8       & 34 \\
        Classes    & 351     & 97  & 160     & 608 \\
        Images     & 448,695 & 124,261   & 206,209 & 779,165 \\
    \end{tabular}
\end{table}

\textbf{Train Categories}:  
\texttt{n02087551} (hound, hound dog), 
\texttt{n02092468} (terrier),  
\texttt{n02120997} (feline, felid),  
\texttt{n02370806} (ungulate, hoofed mammal),
\texttt{n02469914} (primate),  
\texttt{n01726692} (snake, serpent, ophidian),  
\texttt{n01674216} (saurian),
\texttt{n01524359} (passerine, passeriform bird),  
\texttt{n01844917} (aquatic bird),  
\texttt{n04081844} (restraint, constraint),  
\texttt{n03574816} (instrument),  
\texttt{n03800933} (musical instrument, instrument),  
\texttt{n03125870} (craft),
\texttt{n04451818} (tool),  
\texttt{n03414162} (game equipment),
\texttt{n03278248} (electronic equipment),  
\texttt{n03419014} (garment), 
\texttt{n03297735} (establishment), 
\texttt{n02913152} (building, edifice),
\texttt{n04014297} (protective covering, protective cover, protection).

\textbf{Validation Categories}: 
\texttt{n02098550} (sporting dog, gun dog),
\texttt{n03257877} (durables, durable goods, consumer durables),
\texttt{n03405265} (furnishing),
\texttt{n03699975} (machine),  
\texttt{n03738472} (mechanism),
\texttt{n03791235} (motor vehicle, automotive vehicle),

\textbf{Test Categories}: \texttt{n02103406} (working dog),  \texttt{n01473806} (aquatic vertebrate),
\texttt{n02159955} (insect), \texttt{n04531098} (vessel),
\texttt{n03839993} (obstruction, obstructor, obstructer, impediment, impedimenta),
\texttt{n09287968} (geological formation, formation), \texttt{n00020090} (substance),  \texttt{n15046900} (solid).

\section{Extra Experimental Results}
\subsection{Few-shot classification baselines}
We provide baseline results on few-shot classification using 1-nearest neighbor and logistic
regression with either pixel inputs or CNN features. Compared with the baselines, Regular ProtoNet
performs significantly better on all three few-shot classification datasets.

\begin{table}[t]
    \centering
    \resizebox{\textwidth}{!}{
    \begin{small}
    \begin{tabular}{l|c|c|c|c|c}
                 & Omniglot            & \multicolumn{2}{c|}{\textit{mini}ImageNet}& \multicolumn{2}{c}{\textit{tiered}ImageNet}\\
    \cline{2-6}
    Models       & 1-shot              & 1-shot              & 5-shot              & 1-shot              & 5-shot               \\
    \hline
    \hline
    1-NN Pixel   & 40.39 $\pm$ 0.36    & 26.74 $\pm$ 0.48    & 31.43 $\pm$ 0.51    & 26.55 $\pm$ 0.50    & 30.79 $\pm$ 0.53     \\
    1-NN CNN rnd & 59.55 $\pm$ 0.46    & 24.03 $\pm$ 0.38    & 27.54 $\pm$ 0.42    & 25.49 $\pm$ 0.45    & 30.01 $\pm$ 0.47     \\
    1-NN CNN pre & 52.53 $\pm$ 0.51    & 32.90 $\pm$ 0.58    & 40.79 $\pm$ 0.76    & 32.76 $\pm$ 0.66    & 40.26 $\pm$ 0.67     \\
    \hline
    LR Pixel     & 49.15 $\pm$ 0.39    & 24.50 $\pm$ 0.41    & 33.33 $\pm$ 0.68    & 25.70 $\pm$ 0.46    & 36.30 $\pm$ 0.62     \\
    LR CNN rnd   & 57.80 $\pm$ 0.45    & 24.10 $\pm$ 0.50    & 28.40 $\pm$ 0.42    & 26.55 $\pm$ 0.48    & 32.51 $\pm$ 0.52     \\
    LR CNN pre   & 48.49 $\pm$ 0.47    & 30.28 $\pm$ 0.54    & 40.27 $\pm$ 0.59    & 34.52 $\pm$ 0.68    & 43.58 $\pm$ 0.72     \\
    \hline
    ProtoNet     &\tb{94.62 $\pm$ 0.09}&\tb{43.61 $\pm$ 0.27}&\tb{59.08 $\pm$ 0.22}&\tb{46.52 $\pm$ 0.32}& \tb{66.15 $\pm$ 0.34} 
    \end{tabular}
    \end{small}
    }
    \caption{Few-shot learning baseline results using labeled/unlabeled splits. Baselines either
    takes inputs directly from the pixel space or use a CNN to extract features. ``rnd'' denotes
    using a randomly initialized CNN, and ``pre'' denotes using a CNN that is pretrained for
    supervised classification for all training classes.}
    \label{tab:tieredImageNet}
\end{table}

\subsection{Number of unlabeled items}
\begin{figure}
    \centering
    \includegraphics[width=\textwidth]{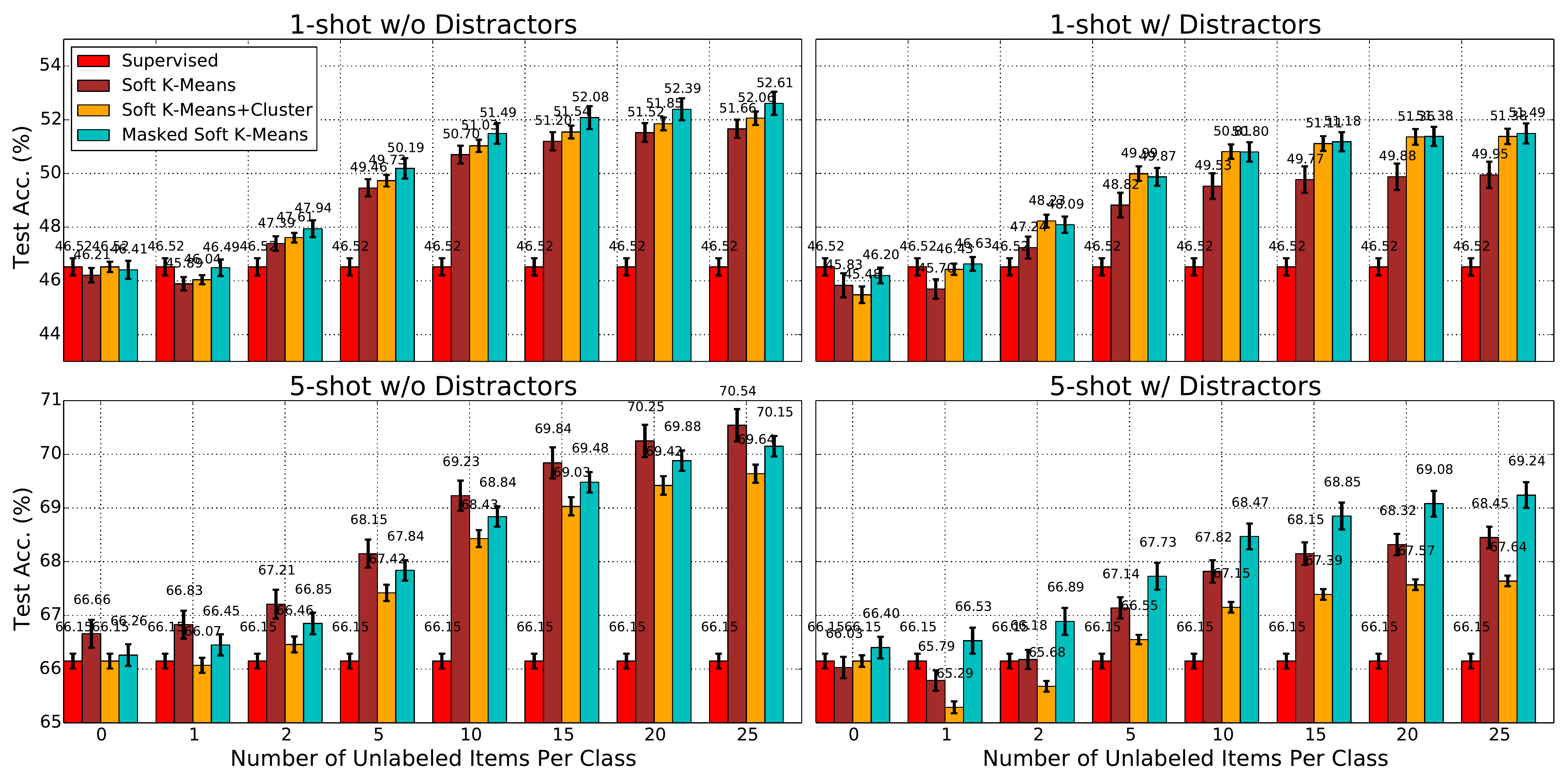}
    \caption{Model Performance on \textit{tiered}ImageNet with different number of unlabeled items during test time. We include test accuracy numbers in this chart.}
    \label{fig:tnet_num_unlabel_text}
\end{figure}

\begin{figure}
    \centering
    \includegraphics[width=0.7\textwidth]{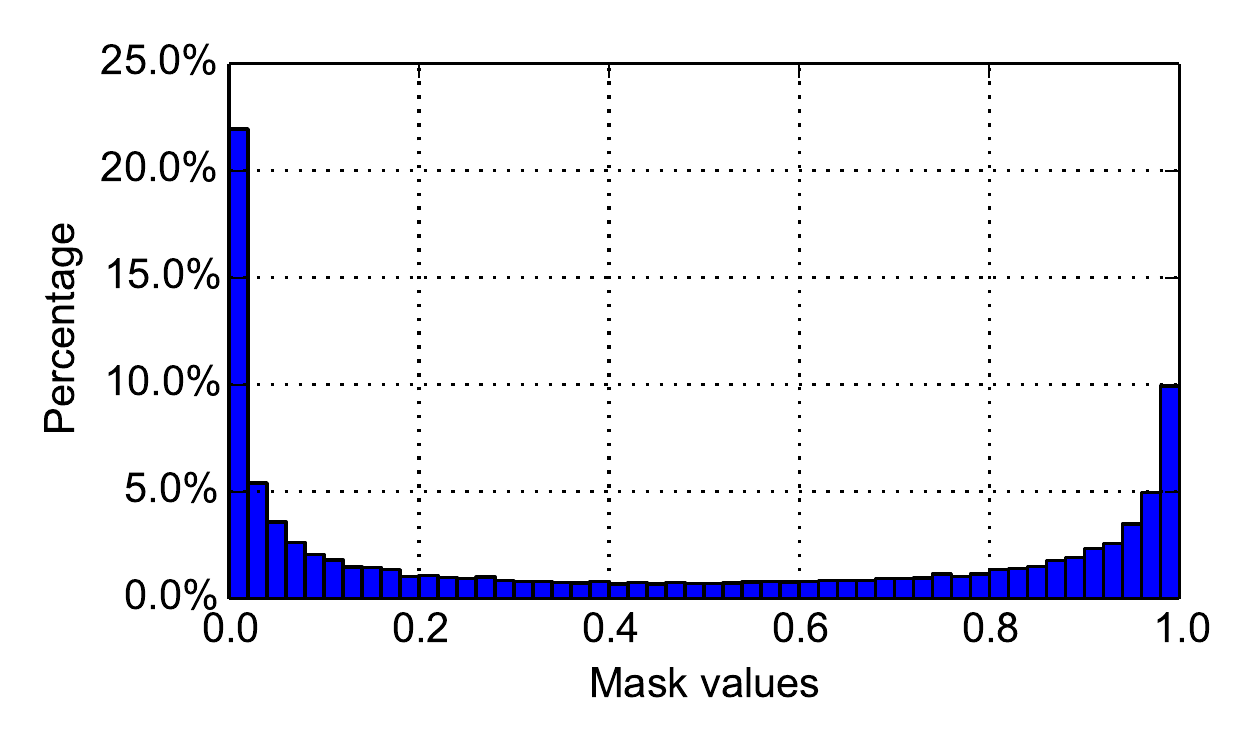}
    \caption{Mask values predicted by masked soft k-means on Omniglot.}
    \label{fig:histo}
\end{figure}

Figure~\ref{fig:tnet_num_unlabel_text} shows test accuracy values with different number of unlabeled items during test time. Figure~\ref{fig:histo} shows our mask output value distribution of the Masked Soft $k$-Means model on Omniglot. The mask values have a bi-modal distribution, corresponding to distractor and non-distractor items.

\section{Hyperparameter Details}
\label{sec:hyperparam}
For Omniglot, we adopted the best hyperparameter settings found for ordinary Prototypical Networks
in \cite{snell2017protonet}. In these settings, the learning rate was set to 1e-3, and cut in half
every $2$K updates starting at update 2K. We trained for a total of 20K updates. For
\textit{mini}Imagenet and \textit{tiered}ImageNet, we trained with a starting learning rate of 1e-3,
which we also decayed. We started the decay after 25K updates, and every 25K updates thereafter we
cut it in half. We trained for a total of 200K updates. We used ADAM \citep{kingma2014adam} for the
optimization of our models. For the MLP used in the Masked Soft $k$-Means model, we use a single
hidden layer with $20$ hidden units with a tanh non-linearity for all $3$ datasets. We did not tune
the hyparameters of this MLP so better performance may be attained with a more rigorous
hyperparameter search.